\documentclass[11pt,letterpaper,logo]{mystyle}
\usepackage{fancyhdr}
\usepackage[utf8]{inputenc}
\usepackage[T1]{fontenc}
\usepackage[numbers]{natbib}
\usepackage{adjustbox}
\usepackage{hyperref}
\usepackage{subcaption}
\usepackage{soul}
\usepackage{multirow}
\usepackage{booktabs}
\usepackage{amsfonts}
\usepackage{nicefrac}
\usepackage{microtype}
\usepackage{xcolor}
\usepackage{colortbl}
\usepackage{enumitem}
\usepackage{amssymb}
\usepackage{amsmath}
\usepackage{wrapfig}
\usepackage{tcolorbox}
\usepackage{placeins}
\usepackage{algorithm}
\usepackage{algpseudocode}
\usepackage{pifont}
\usepackage{bbding}
\usepackage{fontawesome}
\usepackage{array}
\usepackage{xspace}
\usepackage{url}
\tcbuselibrary{skins,breakable}

\definecolor{new_blue}{RGB}{7,14,176}
\definecolor{darkblue}{RGB}{44,52,204}
\definecolor{my_green}{RGB}{0,186,107}
\definecolor{my_red}{RGB}{245,74,69}
\definecolor{hidden-yellow}{RGB}{255,247,200}

\hypersetup{colorlinks=true, citecolor=darkblue, linkcolor=darkblue, urlcolor=darkblue}

\graphicspath{{assets/}{figures/}{logo/}}

\newcommand{\eg}{\textit{e.g.,}\xspace}

\newcommand{\bench}{\textsc{ClawMark}\xspace}
\newcommand{\hm}[1]{\cellcolor{blue!#1}}

\newtcolorbox{findingbox}[1][]{
  colback=LightBlue,
  colframe=LightBlue,
  borderline west={3pt}{0pt}{EvolventAccent},
  boxrule=0pt,
  arc=2pt,
  left=10pt, right=8pt, top=6pt, bottom=6pt,
  before skip=8pt,
  after skip=8pt,
  fontupper=\small\color{EvolventInk},
  #1
}

\fancypagestyle{headstyle}{
  \fancyhead[L]{}
  \fancyhead[C]{\small\textit{ClawMark: A Living-World Benchmark for Multi-Turn Coworker Agents}}
  \fancyhead[R]{}
}

\title{\bench: A Living-World Benchmark for Multi-Turn, Multi-Day, Multimodal Coworker Agents}

\author{%
  \raisebox{-2.5pt}{\includegraphics[height=14pt]{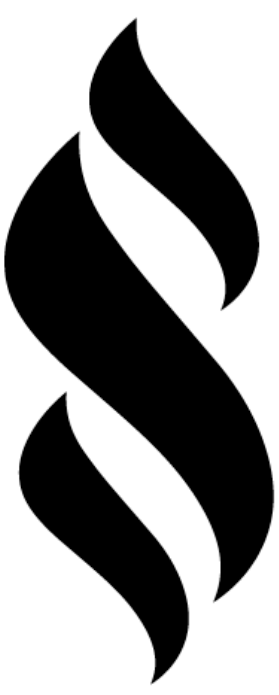}}\hspace{6pt}{\fontsize{12pt}{16pt}\selectfont ClawMark Team}\\
  {\fontsize{9pt}{12pt}\selectfont
  \href{https://github.com/evolvent-ai/ClawMark}{\faGithub~\,github.com/evolvent-ai/ClawMark}\hspace{14pt}%
  \href{https://claw-mark.com/}{\faGlobe~\,claw-mark.com}}%
}

\papertags{%
  \evolventtag{Benchmark}\enspace%
  \evolventtag{Coworker Agents}\enspace%
  \evolventtag{Multimodal}%
}

\begin{document}

% Abstract.

\begin{abstract}
Language-model agents are increasingly used as persistent \emph{coworkers} that assist users across multiple working days. During such workflows, the surrounding environment may change independently of the agent: new emails arrive, calendar entries shift, knowledge-base records are updated, and evidence appears across images, scanned PDFs, audio, video, and spreadsheets. Existing benchmarks do not adequately evaluate this setting because they typically run within a single static episode and remain largely text-centric. We introduce \bench{}, a benchmark for coworker agents built around multi-turn multi-day tasks, a stateful sandboxed service environment whose state evolves between turns, and rule-based verification. The current release contains 100 tasks across 13 professional scenarios, executed against five stateful sandboxed services (filesystem, email, calendar, knowledge base, spreadsheet) and scored by 1{,}537 deterministic Python checkers over post-execution service state; no LLM-as-judge is invoked during scoring. We benchmark seven frontier agent systems. The strongest model reaches 75.8 weighted score, but the best strict Task Success is only 20.0\%, indicating that partial progress is common while complete end-to-end workflow completion remains rare. Turn-level analysis shows that performance drops after the first exogenous environment update, highlighting adaptation to changing state as a key open challenge. We release the benchmark, evaluation harness, and construction pipeline to support reproducible coworker-agent evaluation.
\end{abstract}

\maketitle
\thispagestyle{firststyle}

% §1 Introduction.

\section{Introduction}
\label{sec:intro}

Language-model agents are moving from one-shot task solvers toward persistent \emph{coworkers} that stay with a human across many working days. The surrounding environment keeps evolving while the agent works: new emails arrive, schedules shift, knowledge-base records are updated, and key evidence is distributed across images, scanned PDFs, audio, video, and spreadsheets. Frameworks such as OpenClaw and Claude Code already make this coworker-agent usage pattern increasingly operational. What remains under-measured is a benchmark setting that jointly evaluates agents on these properties, rather than under a single static session.

Existing agent benchmarks still fall short of this setting in three important ways. First, \emph{they evaluate at a time instant rather than across a time interval}: most existing benchmarks score an agent within a single session in which the environment is implicitly assumed not to change between steps. A coworker, by contrast, works across an interval where the environment can evolve between turns: a file read in step 1 may be different in step 2. Second, this assumption is reinforced by a \emph{static-environment design} \cite{yao2024tau,drouin2024workarena,xu2024theagentcompany}: later state changes, when present at all, typically follow from the agent's own actions, rather than from exogenous updates coming from outside the interaction loop. Third, \emph{inputs are text-centric}: although some benchmarks have added images \cite{koh2024visualwebarena,xie2024osworld}, real office work often depends on raw multimodal evidence. The capabilities that distinguish a coworker (persistent state tracking, adaptation to external change, and multimodal evidence integration) therefore remain insufficiently measured.

\bench{} is a benchmark for coworker agents targeting exactly this regime. Each task is a multi-turn workflow spanning multiple in-universe workdays (one turn per working day), executed against five stateful sandboxed services: filesystem, email, calendar, knowledge base, and spreadsheet. By ``stateful sandboxed services'' we mean stateful runtime services running inside the benchmark sandbox (Docker-mounted filesystem, GreenMail SMTP/IMAP, a Notion-compatible knowledge base, a Google-Sheets-compatible spreadsheet, and a Radicale CalDAV server), rather than static logs, cached snapshots, or real third-party production endpoints. Between turns the environment changes independently of the agent, through announced events (which we call \emph{loud events}) and unannounced \emph{silent mutations}, and evidence is delivered untranscribed. Scoring is fully rule-based (\S\ref{sec:design-evaluation}) rather than LLM-as-judge over the agent's prose. The current release contains 100 tasks across 13 professional scenarios.

% T1 --- Structural comparison of agent benchmarks.
% Referenced from sections/01_introduction.tex.

\begin{table}[t]
  \caption{Structural comparison of agent benchmarks along the three axes of \S\ref{sec:intro} plus the scoring choice. \emph{Multimodal}: None / Partial (images only) / Full (audio, video, scanned PDFs, images, spreadsheets; we use ``Full'' here for full coverage of raw non-text office artifacts rather than as a strict modality count). \emph{Multi-Day}: does each task span multiple in-universe days? \emph{Environment}: does external state mutate between turns independently of the agent (Dynamic) or not (Static)? \emph{Verification}: how is task success determined? Among the representative benchmarks compared here, \bench{} is the only one with Multi-Day = Yes, Environment = Dynamic, and Multimodal = Full under rule-based scoring.}
  \label{tab:matrix}
  \centering
  \small
  \setlength{\tabcolsep}{5.0pt}
  \scalebox{0.92}{
  \begin{tabular}{lrrcccc}
    \toprule
    Benchmark        & \# Tasks & \# Scenarios & Multimodal & Multi-Day & Verification    & Environment \\
    \midrule
    WebArena \cite{zhou2023webarena}             & 812 &  5       & None    & No  & Rule-based    & Static \\
    OSWorld \cite{xie2024osworld}               & 369 &  9       & Partial & No  & Rule-based    & Static \\
    Terminal-Bench \cite{merrill2026terminal}  &  89 &  $\sim$6 & None    & No  & LLM-as-judge  & Static \\
    MCPMark \cite{wu2025mcpmark}               & 127 &  5       & None    & No  & Rule-based    & Static \\
    Claw-Eval \cite{ye2026claw}           & 300 &  9       & Full    & No  & Rule + LLM    & Static \\
    ClawsBench \cite{li2026clawsbench}          & 153 & 15       & Partial & No  & LLM-as-judge  & Static \\
    \textbf{\bench{} (Ours)}               & \textbf{100} & \textbf{13} & \textbf{Full} & \textbf{Yes} & \textbf{Rule-based} & \textbf{Dynamic} \\
    \bottomrule
  \end{tabular}
  }
\end{table}

\begin{figure}[t]
  \centering
  \includegraphics[width=\linewidth]{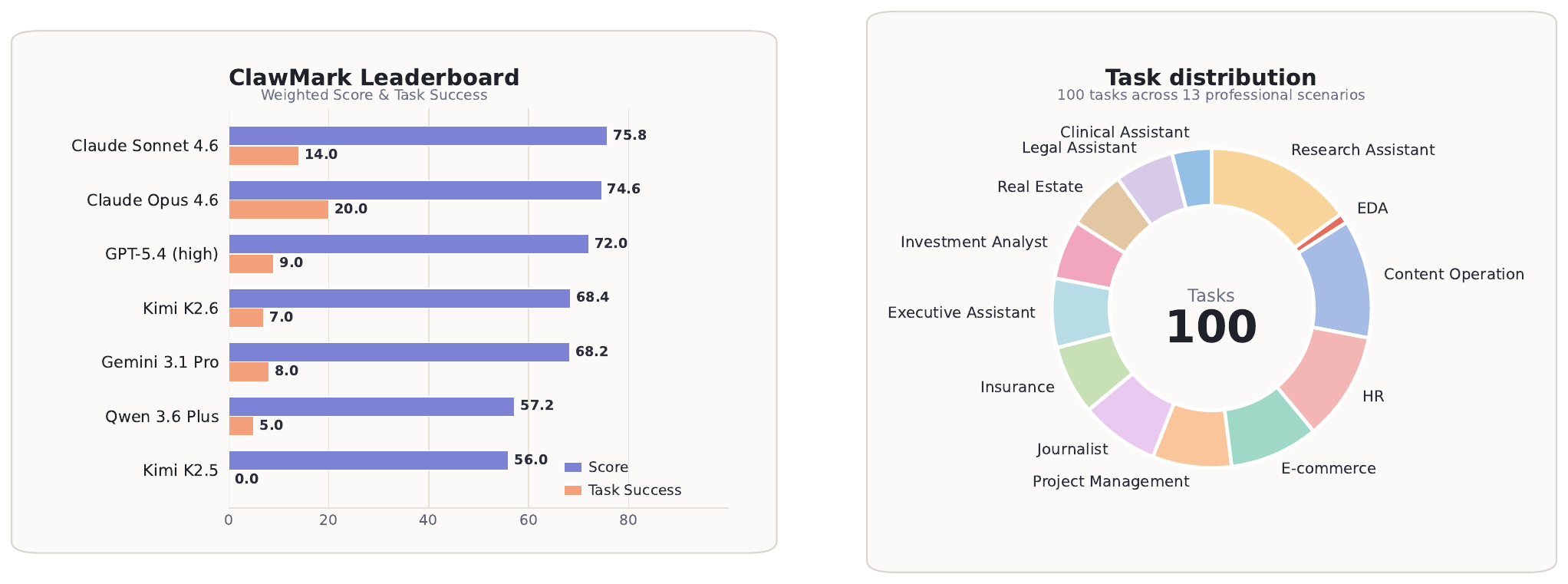}
  \caption{\bench{} results overview. \textbf{Left:} main leaderboard across seven frontier models under the single-run protocol (\S\ref{sec:protocol}); Claude Sonnet 4.6 leads at 75.8 weighted score and the top strict Task Success is 20.0; both metrics leave room to improve. \textbf{Right:} distribution of the 100 tasks across the 13 professional scenarios; the benchmark covers specialised domains including legal assistance, investment analysis, and EDA that prior agent benchmarks have not reached.}
  \label{fig:overview}
\end{figure}

Our contributions are threefold:
\begin{itemize}
  \item We present \bench{}, a benchmark for coworker-agent evaluation that combines multi-turn multi-day tasks, a stateful sandboxed service environment, exogenous between-turn environment changes, and deterministic rule-based scoring in a single executable setting.
  \item We operationalise a \emph{no-LLM-as-judge} scoring protocol: 1{,}537 deterministic Python checkers inspect post-execution service state, and each task is admitted to the released corpus only after two independent re-runs produce bit-identical checker verdicts and diagnostic messages, giving a deterministic alternative to model-judged scoring.
  \item We benchmark seven frontier agent systems on this setting. The top weighted score is 75.8 and the top strict Task Success is 20.0 (these are two different metrics on a 0--100 scale; see \S\ref{sec:design-evaluation}), and both metrics leave room to improve. The per-scenario picture is in Figure~\ref{fig:overview}.
\end{itemize}

Because tasks unfold across multiple turns and external state changes between them, models with similar aggregate scores can follow meaningfully different adaptation trajectories over time. On the 73 tasks with exactly three turns, six of the seven evaluated models drop on Day 2, while the Claude Sonnet 4.6--GPT-5.4 gap narrows from $+6.5$ percentage points on Day 1 to $+4.0$ percentage points on Day 3 (\S\ref{sec:exp-trajectory}). We release the benchmark, the evaluation harness, and the construction pipeline to support reproducible evaluation of the next generation of coworker agents.

% F1 --- Results overview: leaderboard bar chart (left) + task distribution
% donut (right) rendered in a single unified figure (assets/results_overview.pdf).
% Referenced from sections/01_introduction.tex.

% §2 Related work.

\section{Related work}
\label{sec:related}

\subsection{Agent benchmarks}
\label{sec:related-benchmarks}

Table~\ref{tab:matrix} positions \bench{} against representative agent benchmarks. Realistic web and computer-use benchmarks such as WebArena \cite{zhou2023webarena}, Mind2Web \cite{deng2023mind2web}, VisualWebArena \cite{koh2024visualwebarena}, and OSWorld \cite{xie2024osworld} establish strong single-episode evaluation settings. Related benchmarks extend tool coverage or execution domains, for example MCPMark \cite{wu2025mcpmark}, MCP-Bench \cite{wang2025mcp}, SWE-bench \cite{jimenez2023swe}, AgentBench \cite{liu2023agentbench}, GAIA \cite{mialon2023gaia}, and Terminal-Bench \cite{merrill2026terminal}, but still largely evaluate progress within a fixed episode.

Multi-turn benchmarks such as tau-bench \cite{yao2024tau}, WorkArena \cite{drouin2024workarena}, and TheAgentCompany \cite{xu2024theagentcompany} move beyond single-turn execution, yet later state changes still arise primarily from the interaction itself rather than from exogenous changes between workdays. Concurrent Claw-$\ast$ benchmarks target adjacent gaps: Claw-Eval \cite{ye2026claw} adds trajectory-aware scoring, ClawsBench \cite{li2026clawsbench} trades reproducibility for ecological validity on live websites, and ClawArena \cite{ji2026clawarena} studies evolving information streams through perception-level questioning. \bench{} differs by combining multi-day tasks, exogenous between-turn mutation, raw multimodal evidence, and fully rule-based verification under deterministic checker re-runs (\S\ref{sec:construction-pipeline}).

\subsection{LLM agents}
\label{sec:related-agents}

Rapid progress in LLM-based agent systems has enabled reliable multi-step tool use, code execution, and cross-service orchestration. Research frameworks such as SWE-agent \cite{yang2024swe}, AutoGen \cite{wu2024autogen}, MetaGPT \cite{hong2023metagpt}, and CAMEL \cite{li2023camel} study how language agents can act through tools, roles, and multi-agent interaction. Product and open-source scaffolds such as OpenClaw, Claude Code, Cursor, AutoGPT, and AgentGPT make similar capabilities operational against filesystems, shells, browsers, and external APIs.

Most of these systems are still evaluated in episodic settings where the environment resets between tasks. \bench{} targets the complementary coworker-agent regime, where the agent persists across in-universe working days, refreshes an independently mutating external state at the start of each turn, and operates over raw multimodal evidence. \bench{} is \emph{framework-compatible}: the tool schema for our five services is harness-agnostic by design, and any agent framework that implements it can be scored. We report all seven models in the main table under a single harness (OpenClaw, \S\ref{sec:exp-setup}) to isolate model-side differences. This claim concerns interface design only; we do not claim empirical equivalence across different agent frameworks.

% §3 ClawMark.

\section{\bench}
\label{sec:design}

\subsection{Overview}
\label{sec:design-overview}

Each \bench{} task simulates a realistic office workflow that a coworker agent must carry out alongside its human user across multiple working days. A task is composed of two to six \emph{turns}, where one turn corresponds to one in-universe working day, delivered as a wake-up message and executed against five stateful sandboxed services: filesystem, email, calendar, knowledge base, and spreadsheet. Between turns the framework mutates external state independently of the agent, as announced events (which we call \emph{loud events}) delivered in the wake-up message, or as \emph{silent mutations} that appear in the services without notification, so a competent coworker must refresh external state at each turn rather than act on cached assumptions from the previous day. Raw multimodal artifacts (photos, audio, scanned PDFs, video, spreadsheets) are first-class evidence, delivered without pre-transcription.

\begin{figure}[t]
  \centering
  \includegraphics[width=\linewidth]{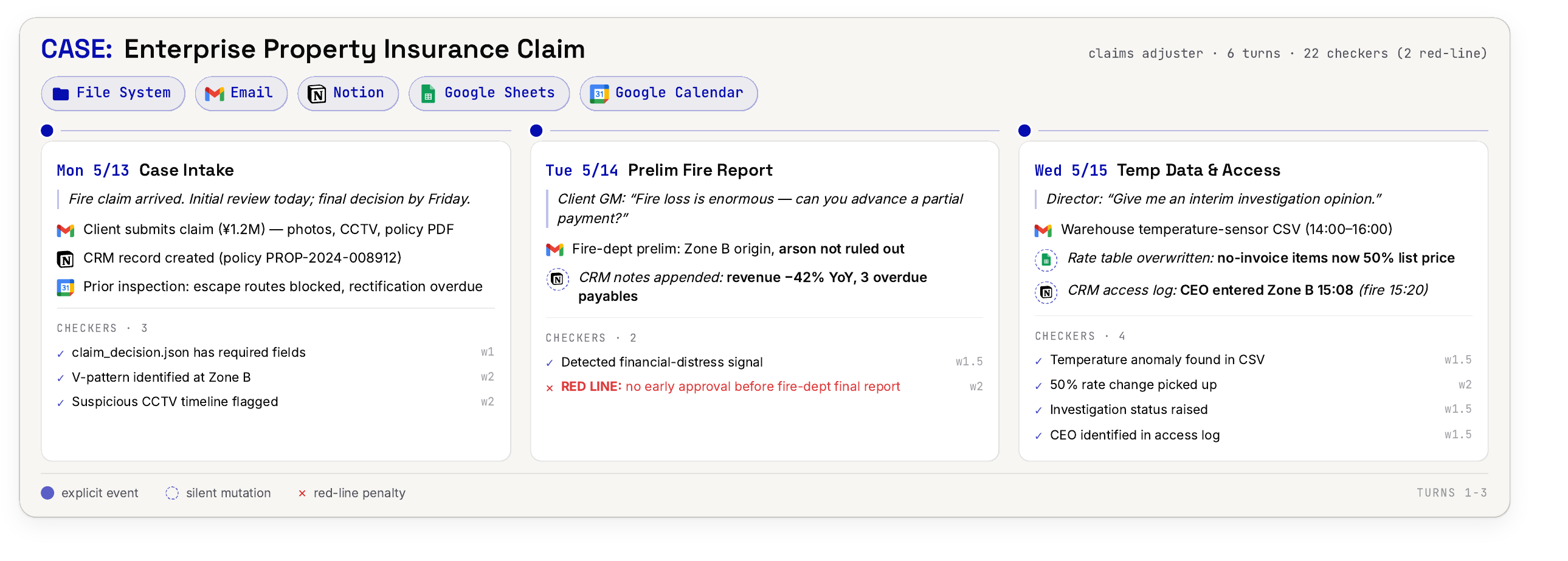}
  \caption{Anatomy of a \bench{} task. Example: \texttt{insurance\_task5} (Enterprise Property Insurance Claim), a six-turn adjudication of a \textyen1.2\,M fire-damage claim with 22 weighted checkers across five backends; turns 1--3 are shown here; the remaining three turns follow the same template (wake-up prompt, loud/silent events, per-turn checkers). Each card is one in-universe working day. Coloured pills list the backends the turn touches. Italicised blocks are the turn-entry prompts. Solid dots denote \emph{explicit} events delivered in the prompt; dashed dots denote \emph{silent} mutations injected between turns without a notification (on Wed 5/15, the warehouse sensor log, the overwritten rate table, and the access-log entry all appear without being requested by the agent). Checker rows show per-turn rubric items with weights (\texttt{w1}, \texttt{w1.5}, \texttt{w2}); \textbf{red-line} items (in red) are high-weight hard constraints implemented as deterministic checker failures inside the same weighted rubric (here the day-2 red-line forbids approving or rejecting the claim before the fire-department final report arrives). Scoring is fully rule-based: each checker is a deterministic Python function that inspects post-turn service state.}
  \label{fig:task-card}
\end{figure}

Scoring is fully rule-based: every task ships with 6--29 weighted Python checkers that query the post-turn state of the sandboxed services, and the task score is their weight-normalised pass rate on $[0, 1]$. The complete rule-based-scoring commitment, including the no-LLM-as-judge guarantee, is defined in \S\ref{sec:design-evaluation}. Figure~\ref{fig:task-card} illustrates these elements on \texttt{insurance\_task5}, a six-turn enterprise fire-claim adjudication.

% F2 --- Task anatomy card for insurance_task5.
% Rendered from scripts/fig02_task_card.html via headless Chrome:
%   chrome --headless --print-to-pdf=assets/task_card.pdf fig02_task_card.html
% Referenced from sections/03_design.tex.

The current release contains 100 tasks across 13 professional scenarios and 87 distinct in-task roles, running against the five sandboxed services with 1{,}072 raw multimodal artifacts (PDFs, images, audio, video, spreadsheets) and scored by 1{,}537 deterministic Python checkers of which 55 are red-line constraints. Tasks range from two to six turns (mean 3.6) and 6 to 29 checkers (mean 15.4). The corpus was produced by a task-first construction pipeline with multi-round human and agent-assisted review, described in detail in Section~\ref{sec:construction}.

\subsection{Evaluation}
\label{sec:design-evaluation}

At evaluation time the framework runs each checker against the post-turn state of the sandboxed services (not a cached snapshot), and reports both the weight-normalised task score and a stricter aggregate \emph{Task Success} metric. Throughout the paper, both are reported on a 0--100 percentage scale for readability.
\begin{align}
  \mathrm{score}(m, \tau) \;&=\; \frac{\sum_{c \in \mathcal{C}(\tau)} w_c \cdot \mathbf{1}[\mathrm{pass}_c(m, \tau)]}{\sum_{c \in \mathcal{C}(\tau)} w_c} \; \in\; [0, 1]\, ,
  \label{eq:score}\\
  \mathrm{Succ}(m) \;&=\; \frac{100}{|\mathcal{T}|} \sum_{\tau \in \mathcal{T}} \mathbf{1}\!\left[\forall c \in \mathcal{C}(\tau),\, \mathrm{pass}_c(m, \tau) \right] \, \in [0, 100] \, ,
  \label{eq:task-success}
\end{align}
where $\mathcal{C}(\tau)$ denotes the checker set of task $\tau$ with per-checker weights $w_c$, $\mathcal{T}$ is the full 100-task corpus, and $\mathrm{pass}_c(m, \tau) \in \{0, 1\}$ is the deterministic pass/fail verdict that checker $c$ produces against the post-turn service state left by model $m$ on task $\tau$. Each of the 1{,}537 checkers falls into one of four categories: filesystem / artifact inspection, external-backend state queries, email state queries, and numeric-tolerance or semantic-equivalence checks. \textbf{No LLM judge is invoked at any point during scoring}, whether during pass/fail decisions, tolerance checks, or aggregation; every verdict is a deterministic function of post-turn service state. A subset of 55 checkers (3.6\%) are designated \emph{red-line} constraints capturing compliance-sensitive actions a coworker must never take, and fall into four classes: premature-decision, compliance-bypass, data-exfiltration, and irreversible-write. In implementation, these red-line constraints are ordinary checker entries with fixed high weights $w_{\mathrm{red}}$ inside the same rubric, so a task with every non-red checker passing can still score substantially below $1.0$ if a red-line checker fails. Per-scenario red-line counts are reported in Table~\ref{tab:scenarios}.

\paragraph{Why report both weighted Score and strict Task Success.} The two metrics answer different questions. Eq.~\eqref{eq:score} is a continuous signal that rewards partial progress and is appropriate for leaderboard ordering when rubrics vary in length (tasks in \bench{} have 6--29 checkers). Eq.~\eqref{eq:task-success} is a binary all-or-nothing signal that asks whether an agent completed the \emph{entire} workflow a coworker was asked to perform, which is the deployment-relevant question in a professional setting. Because Task Success requires every checker to pass, it is sensitive to single-item rubric brittleness, long-tail checker dependencies, and the presence of red-line constraints (red-lines count as ordinary pass/fail items inside this metric, since the all-or-nothing aggregation does not reference per-checker weights). We therefore always report Task Success alongside weighted score rather than instead of it.

\subsection{Design principles}
\label{sec:design-principles}

Relative to prior agent benchmarks, \bench{} makes three design commitments. \textbf{Multi-turn timelines}: each task spans two to six in-universe working days (one day per turn) with clock advancement between turns, so the agent must sustain progress across day boundaries rather than emit a single-shot trajectory. \textbf{Dynamic environment}: between-turn mutation is injected in two layers. An \texttt{inject/stage\{N\}/} directory (legacy field name; one entry per turn) drops new files into the workspace, and turn-entry service-side Python appends email, rewrites spreadsheet rows, edits knowledge-base entries, and shifts calendar events. The agent therefore must refresh external state at the start of each turn rather than act on a day-1 mental model. \textbf{Full multimodal evidence}: raw artifacts are delivered without pre-transcription, and models must parse them with their own tools (\texttt{whisper}, \texttt{ffmpeg}, \texttt{PyMuPDF}, etc.).

At the implementation level, every task is fully specified by a single \texttt{task.py} together with per-turn inject layers and supporting artifacts, and runs against the same five services (a Docker-mounted filesystem, GreenMail for SMTP/IMAP, a Notion-compatible knowledge base, a Google-Sheets-compatible spreadsheet, and a Radicale CalDAV server) inside an isolated \texttt{docker-compose} group. Appendix~\ref{app:task-diagram} provides a compact implementation-level view of how these task files are parsed into runtime objects and checked by the rule-based evaluation pipeline.

Section~\ref{sec:construction} details how this corpus is produced: task distribution (\S\ref{sec:construction-distribution}) and the task-first construction pipeline and release gate (\S\ref{sec:construction-pipeline}).

% §4 Benchmark Construction.

\section{Benchmark construction}
\label{sec:construction}

\subsection{Task distribution}
\label{sec:construction-distribution}

The 13 scenarios cover both general office roles (executive assistant, HR, content operation, e-commerce, journalist, project management, real estate, research assistant) and specialised professional domains that most existing agent benchmarks have not reached (clinical assistant, insurance, legal assistant, investment analyst, electronic design automation). The 87 in-task roles are substantive rather than cosmetic: the clinical assistant scenario alone includes a pharmacist assistant, a surgical scheduler, a charge nurse, and a chronic-disease clinic assistant, each with its own rubric. Per-scenario task, role, turn, and checker counts, together with red-line counts and dynamic-environment composition statistics (mean per-task shares of silent vs.\ loud between-turn changes), are reported in Table~\ref{tab:scenarios}.

% T2 --- Scenario composition of ClawMark.
% All columns derived from /Users/moonshot/Desktop/project/ClawMark/tasks:
%   # Tasks   = count of task directories under tasks/{scenario}/
%   # Roles   = distinct METADATA["role"] strings within the scenario
%   Turns     = mean turns executed per task (from result.json stages array)
%   Checkers  = mean count of unique checker IDs per task
%   Red-line  = total checker IDs matching /red[_-]?line/ in task.py source
%   Silent/Loud % = mean per-task share of explicitly annotated environment-injection
%                   events (turn1+) classified as silent (un-announced backend mutations)
%                   or loud (announced new arrivals / notifications). Per-task counts
%                   were derived by reading each task.py against author-supplied
%                   `# Silent` / `# Loud` comments using LLM-assisted classification.
% Referenced from sections/03_design.tex / sections/04_construction.tex.

\begin{table}[t]
  \caption{Scenario composition of \bench{} (100 tasks, 87 distinct in-task roles, 55 red-line checkers). \emph{\# Roles} counts distinct \texttt{METADATA["role"]} strings; \emph{Turns} and \emph{Checkers} are mean values per task; \emph{Red-line} is the total red-line checker count in the scenario (\S\ref{sec:design-principles}). \emph{Silent \%} and \emph{Loud \%} report the mean per-task share of annotated between-turn injection events classified as \emph{silent} (un-announced) or \emph{loud} (announced). Silent/loud annotations are descriptive metadata used only for corpus characterisation, not for scoring; the classification follows author-supplied \texttt{\# Silent} / \texttt{\# Loud} comments in each \texttt{task.py} and is sensitive to borderline cases at the per-scenario level (further discussion in \S\ref{sec:limits}).}
  \label{tab:scenarios}
  \centering
  \small
  \setlength{\tabcolsep}{5pt}
  \begin{tabular}{lrrrrrrr}
    \toprule
    Scenario & \# Tasks & \# Roles & Turns & Checkers & Red-line & Silent \% & Loud \% \\
    \midrule
    Clinical Assistant    &   4 &  4 & 3.5 & 17.5 & 15 & 58.5 & 41.5 \\
    Content Operation     &  12 & 11 & 3.3 & 13.5 &  3 & 49.9 & 50.1 \\
    E-commerce            &   9 &  5 & 3.1 & 15.0 &  0 & 52.1 & 47.9 \\
    EDA                   &   1 &  1 & 2.0 & 18.0 &  0 & 50.0 & 50.0 \\
    Executive Assistant   &   7 &  7 & 3.7 & 17.7 &  0 & 47.8 & 52.2 \\
    HR                    &  11 & 11 & 3.7 & 20.6 &  3 & 30.2 & 69.8 \\
    Insurance             &   7 &  7 & 5.1 & 13.0 & 14 & 66.9 & 33.1 \\
    Investment Analyst    &   6 &  6 & 4.0 & 17.7 &  0 & 50.0 & 50.0 \\
    Journalist            &   8 &  8 & 2.9 & 12.5 &  1 & 28.2 & 71.8 \\
    Legal Assistant       &   6 &  1 & 3.0 & 13.0 &  0 & 46.0 & 54.0 \\
    Project Management    &   8 &  7 & 3.1 & 13.1 &  6 & 56.2 & 43.8 \\
    Real Estate           &   6 &  6 & 3.5 & 18.5 &  2 & 53.7 & 46.3 \\
    Research Assistant    &  15 & 13 & 3.9 & 14.0 & 11 & 40.9 & 59.1 \\
    \midrule
    \textbf{Total}        & \textbf{100} & \textbf{87} & \textbf{3.6} & \textbf{15.4} & \textbf{55} & \textbf{46.7} & \textbf{53.3} \\
    \bottomrule
  \end{tabular}
\end{table}

\subsection{Construction pipeline}
\label{sec:construction-pipeline}

\begin{wrapfigure}{r}{0.40\linewidth}
  \vspace{-10pt}
  \centering
  \includegraphics[width=\linewidth,keepaspectratio]{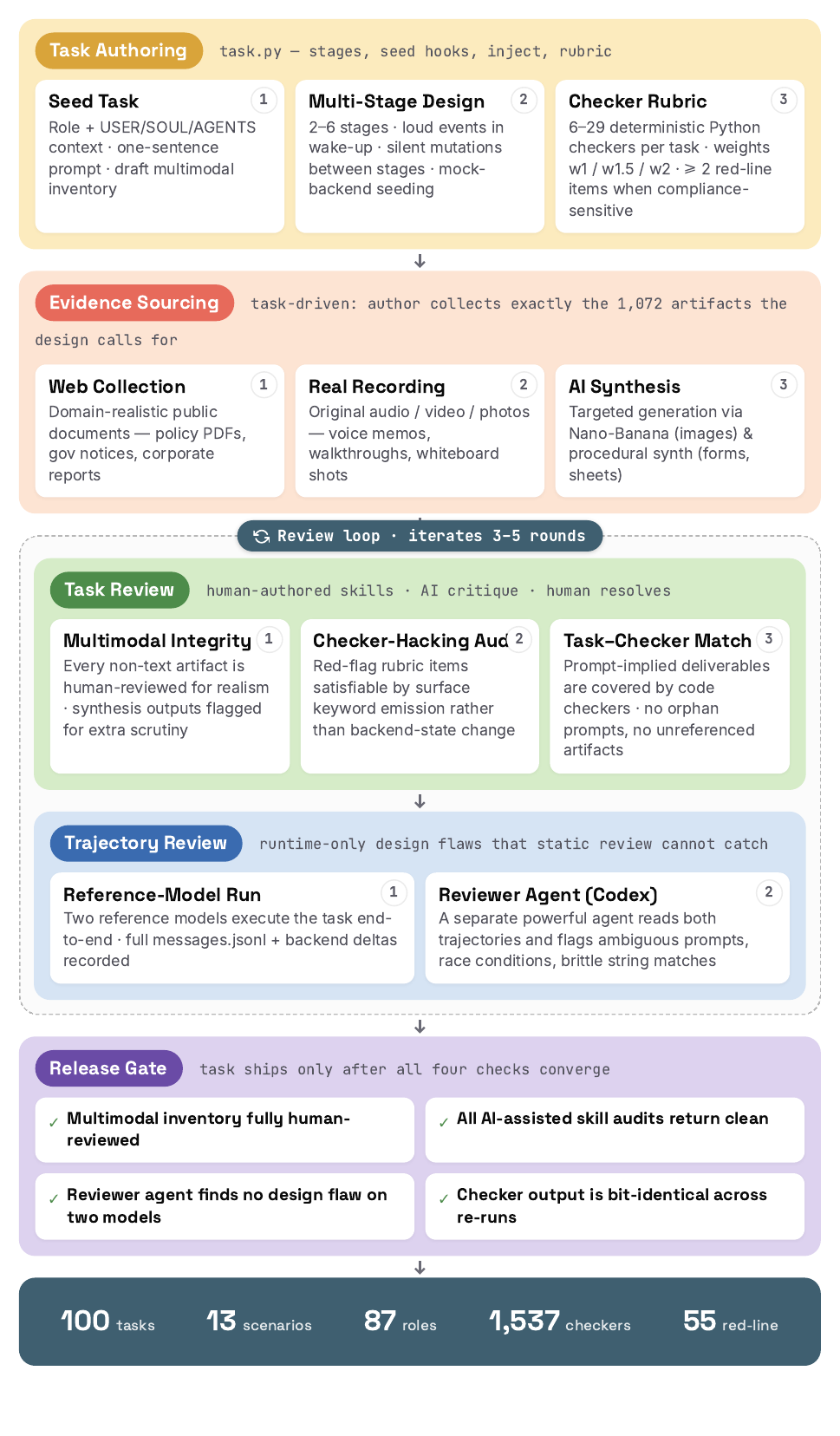}
  \caption{\bench{} construction pipeline. Four phases: task authoring, task-driven evidence sourcing, a review loop (task review + trajectory review) that iterates 3--5 rounds per task, and a release gate. A task enters the release corpus only when all four release-gate conditions hold simultaneously.}
  \label{fig:pipeline}
  \vspace{-10pt}
\end{wrapfigure}

Producing 100 tasks across 13 heterogeneous professional scenarios while preserving multimodal authenticity and deterministic checker correctness presents a nontrivial authoring challenge. Text-only generation pipelines are insufficient on their own in this setting: the evaluation signal depends on artifacts that must look, sound, and parse like their real-world counterparts, and every checker must run against a stateful sandboxed service with bit-identical output across re-runs. We therefore adopt a \emph{task-first} pipeline with four phases (task authoring, task-driven evidence sourcing, a review loop that repeats 3--5 rounds per task, and a release gate), summarised in Figure~\ref{fig:pipeline}. Pipeline phases are distinct from in-task turns: each phase below is an authoring step, while a turn is one in-universe working day inside an executed task.

\paragraph{Phase 1: Task authoring.} The pipeline starts with task design, because the specification determines the multimodal inventory rather than the other way round. Each author writes a single \texttt{task.py} containing turn definitions, service seed hooks, between-turn injections (loud events and silent mutations), and a weighted checker rubric. Three invariants guide this step: every silent mutation is tied to at least one checker; every cross-modal contradiction spans at least two modalities; and every red-line is expressed as a deterministic state check rather than prose matching. The output is a concrete artifact list for the scenario, handed off to Phase~2.

\paragraph{Phase 2: Evidence sourcing.} Each required artifact is produced through one of three channels with a provenance tag: \emph{web collection} of domain-realistic public documents (policy PDFs, government notices, corporate reports); \emph{original recording} of audio / video / photographs (voice memos, walkthrough videos, whiteboard photos); and \emph{targeted AI synthesis} (\eg{} Nano-Banana for photographs, procedural generators for forms and spreadsheets). The authoring-first order matters: the opposite direction (collect a corpus first, then craft tasks around what happens to be available) produces information-dense but purpose-ambiguous artifacts whose checker coverage is incidental rather than by design.

% F5 --- Construction pipeline.
% Rendered from scripts/fig05_pipeline.html via headless Chrome.
% Referenced from sections/04_construction.tex. wrapfigure on the right
% so §4.2 prose flows around it on the left.

\paragraph{Phase 3: Review loop (3--5 rounds).} Every task alternates between \emph{task review} and \emph{trajectory review}. Task review combines human artifact inspection with three AI audits: \emph{multimodal integrity}, \emph{checker-hacking}, and \emph{task--checker correspondence}. Trajectory review runs two reference models end-to-end and asks an independent Codex-class reviewer agent to flag runtime-only design flaws such as ambiguous turn prompts, inject--checker races, under-specified deliverable schemas, and brittle string matches. Findings return to the author for revision, and the loop iterates 3 to 5 times per task.

\paragraph{Phase 4: Release gate.} A task enters the released corpus only when four conditions hold simultaneously: (i) human sign-off on every multimodal artifact; (ii) clean results from all three task-review audits; (iii) no design-flaw finding from the reviewer agent on trajectories from two distinct reference models; and (iv) \emph{bit-identical} checker verdicts and detail messages across two independent re-runs against the same frozen service state. Condition~(iv) is the operational guarantee behind the no-LLM-as-judge claim; tasks that fail it more than twice are redesigned or dropped.

% §5 Experiments.

\section{Experiments}
\label{sec:exp}

We evaluate seven frontier models end-to-end on \bench{}, covering five proprietary models and two open-source models. This section describes the experimental setup (\S\ref{sec:exp-setup}) and presents the main leaderboard together with the per-scenario breakdown (\S\ref{sec:exp-main}). Turn-level trajectory analysis and failure taxonomy follow in \S\ref{sec:analysis}; case studies are deferred to Appendix~\ref{app:case-studies}.

\subsection{Experimental Setup}
\label{sec:exp-setup}
\label{sec:protocol}

\paragraph{Models.} We evaluate five proprietary models (\textbf{Claude Sonnet 4.6}, \textbf{Claude Opus 4.6}, \textbf{GPT-5.4 (high)}, \textbf{Gemini 3.1 Pro Preview}, and \textbf{Qwen 3.6 Plus}) and two open-source models (\textbf{Kimi K2.5} and \textbf{Kimi K2.6}). The Kimi K2.5 result reported here is from the public Infinigence-hosted endpoint; an internal instance-tuned Kimi K2.5 variant exists but is not reported in the main table to avoid mixing a public and a private baseline.

\paragraph{Framework and infrastructure.} Every model runs under a single agent framework, \textbf{OpenClaw}, with identical tool schemas across models; no per-model prompt engineering is performed. For the Kimi-series models, we apply the upstream fix for incorrect tool-call identifier sanitisation in OpenClaw\footnote{\url{https://github.com/openclaw/openclaw/issues/62319}}. Each task executes inside an isolated \texttt{docker-compose} group comprising the agent container, GreenMail for SMTP/IMAP, a Notion-compatible knowledge base, a Google-Sheets-compatible spreadsheet, and a Radicale CalDAV server. Containers are torn down between tasks, so runs do not share state.

\paragraph{Evaluation settings and metric.} All models use provider-default inference parameters, with \emph{extended thinking} enabled where supported (Claude, GPT-5.4, Gemini) and prompt caching enabled where providers offer it. Weighted score and Task Success follow \S\ref{sec:design-evaluation}; both are reported on a 0--100 scale throughout this section. We additionally report wall-clock time for a full benchmark sweep and total input/output tokens. \textbf{Main-table results are from a single full sweep per model.} We do not report run-to-run variance in this release; rankings among models that fall within a small weighted-score band should be read with that caveat in mind.

\paragraph{Cost normalisation.} For the efficiency view in \S\ref{sec:exp-main}, we treat total tool calls and total (input + output) tokens as compute-side proxies rather than as direct dollar-cost signals: API unit pricing differs by provider and changes month to month, so a frozen cost-per-token table would misrepresent at least one model by the time of camera-ready. Readers who need a monetary bound can recombine the per-model token counts in Table~\ref{tab:leaderboard} with the provider's published rate at read time.

\subsection{Main Results}
\label{sec:exp-main}

Table~\ref{tab:leaderboard} reports the overall leaderboard across seven models and 100 tasks. By weighted score, Claude Sonnet 4.6 (75.8), Claude Opus 4.6 (74.6), and GPT-5.4 (72.0) cluster within a 3.8\,pp band; because each model is evaluated with a single full sweep, rankings among models within this narrow band should be interpreted cautiously, and the larger gap to GPT-5.4 should likewise be read in the absence of repeated sweeps to quantify run-to-run variance. Under the stricter Task Success metric of Eq.~\eqref{eq:task-success}, the ordering changes: Claude Opus 4.6 leads at 20.0, followed by Sonnet 4.6 at 14.0 and GPT-5.4 at 9.0, and fully correct end-to-end completion is much rarer than partial progress.

% T4 --- Main leaderboard.
% Referenced from sections/05_experiments.tex.

\begin{table}[t]
  \caption{Main results on \bench{} (single-sweep). \emph{Score} is the mean of Eq.~\eqref{eq:score} across the 100 tasks, reported on a 0--100 scale. \emph{Task Success} follows Eq.~\eqref{eq:task-success}. \emph{Red-line fail} is the fraction of red-line checker evaluations the model fails, aggregated over the 55 red-line checkers (distributed across the 8 scenarios that carry any; see Table~\ref{tab:scenarios}). \emph{Wall time} is the total wall-clock for a full single-run sweep. \emph{Input} merges \texttt{input + cacheRead + cacheWrite} tokens so providers are comparable regardless of prompt-cache use. \emph{Tool calls} is the total across 100 tasks (mean 48--71 per task). Main-table results are from a single full sweep per model; run-to-run variance is not reported in this release, so ranking claims among models that fall within a small weighted-score band should be read as tentative.}
  \label{tab:leaderboard}
  \centering
  \scriptsize
  \setlength{\tabcolsep}{5pt}
  \begin{tabular}{lrrrrrrr}
    \toprule
    Model                         & Score & Task Success & Red-line fail & Wall time & Input tok.\ & Output tok.\ & Tool calls \\
    \midrule
    Claude Sonnet 4.6             & \textbf{75.8} & 14.0          &  3.6\% & 22.3\,h & 257.8\,M & 2.57\,M & 5{,}736 \\
    Claude Opus 4.6               & 74.6          & \textbf{20.0} &  5.5\% & 22.6\,h & 266.7\,M & 2.02\,M & 6{,}112 \\
    GPT-5.4 (high)                & 72.0          & 9.0           &  3.6\% & 26.1\,h & 231.5\,M & 2.93\,M & 7{,}052 \\
    Kimi K2.6                     & 68.4          & 7.0           &  7.3\% & 22.6\,h & 226.3\,M & 2.30\,M & 6{,}026 \\
    Gemini 3.1 Pro Preview        & 68.2          & 8.0           &  3.6\% & 18.9\,h & 338.8\,M & 1.77\,M & 5{,}877 \\
    Qwen 3.6 Plus                 & 57.2          & 5.0           & 14.5\% & 33.3\,h & 315.1\,M & 4.56\,M & 6{,}119 \\
    Kimi K2.5                     & 56.0          & 0.0           &  9.1\% & 22.8\,h & 214.0\,M & 1.47\,M & 4{,}776 \\
    \bottomrule
  \end{tabular}
\end{table}

\paragraph{Both metrics leave room for improvement.} No model exceeds 75.8 weighted score overall. Excluding the single-task EDA case, the highest per-scenario score is Claude Opus 4.6 at 92.6 on real estate, still leaving visible headroom. The stricter Task Success metric makes the gap clearer: even the strongest model fully solves only 20.0\% of tasks, Sonnet 4.6 fully solves 14.0\%, GPT-5.4 fully solves 9.0\%, and Kimi K2.5 fully solves 0.0\%. On the hardest scenario, project management, every model scores below 44.0, and the mean weighted score across seven models is about 35.1. A partially correct trajectory that misses a single silent mutation, an incomplete backend writeback, or a turn-specific rubric item still forfeits Task Success credit, which is consistent with \bench{}'s task design.

\paragraph{Per-scenario best is distributed across four models, not one.} Table~\ref{tab:scenario-heatmap} shows outright-best-by-scenario splits among four models: Claude Sonnet 4.6 on clinical assistant, e-commerce, HR, legal, and research assistant (five); Claude Opus 4.6 on content operation, insurance, journalist, project management, and real estate (five); GPT-5.4 on executive assistant; and Gemini 3.1 Pro Preview on investment analyst. The two Anthropic models tie on the single EDA task (100.0 each). Below third place the ranking reshuffles more sharply: Kimi K2.6 is competitive with Gemini on investment analyst (82.1 vs.\ 82.9), but trails it sharply on EDA (8.7 vs.\ 91.3) due to a single vision-dependent task it routed incorrectly. Coworker-agent evaluation therefore does not collapse to a single frontier-model ordering, and specialisation-driven scenarios are where models most separate: EDA separates Gemini from the Kimi family (with the caveat that EDA contains a single task and should be read as a case-level result rather than a stable scenario-level trend), project management is uniformly difficult for all models, and clinical / insurance / research-assistant are where red-line-heavy rubrics (Table~\ref{tab:scenarios}) reward compliance-aware trajectories.

\paragraph{No monotone relationship between score and tool or token consumption.} Three per-model point comparisons illustrate: +23\% tool calls at $-3.8$\,pp score (GPT-5.4 vs.\ Sonnet), +31\% input tokens at $-7.6$\,pp (Gemini vs.\ Sonnet), and $1.8\times$ output tokens at $-18.6$\,pp (Qwen vs.\ Sonnet). On \emph{score per thousand tool calls} (a compute-side proxy for action efficiency; see Cost normalisation in \S\ref{sec:exp-setup}), Sonnet 4.6 leads at 13.2, followed by Opus 4.6 (12.2), Kimi K2.5 (11.7), Gemini (11.6), Kimi K2.6 (11.3), GPT-5.4 (10.2), and Qwen (9.3). The two top-scoring models are also the two most tool-efficient, so under this proxy score and efficiency move together rather than trade off.

% T5 --- Per-scenario score heatmap (7 models x 13 scenarios).
% Referenced from sections/05_experiments.tex.
% Cell shading via \hm{N} macro (N = score * 100), defined in neurips_2026.tex.

\begin{table}[t]
  \caption{Per-scenario score on a 0--100 scale. Bold marks the per-scenario best. Cell shading is proportional to score (darker = higher). The per-scenario best is distributed across four models (Claude Sonnet 4.6, Claude Opus 4.6, GPT-5.4, and Gemini 3.1 Pro Preview); the two Anthropic models tie on EDA at 100.0. EDA contains a single task (Table~\ref{tab:scenarios}), so its row should be read as a case-level result rather than a stable scenario-level trend.}
  \label{tab:scenario-heatmap}
  \centering
  \footnotesize
  \setlength{\tabcolsep}{3.3pt}
  \begin{tabular}{lccccccc}
    \toprule
    Scenario                 & Sonnet 4.6 & Opus 4.6 & GPT-5.4 & Kimi K2.6 & Gemini 3.1 & Qwen 3.6 & Kimi K2.5 \\
    \midrule
    Clinical Assistant       & \hm{70}\textbf{92.3} & \hm{67}88.3 & \hm{66}86.9 & \hm{66}87.4 & \hm{55}73.2 & \hm{59}77.9 & \hm{53}71.0 \\
    Content Operation        & \hm{50}66.2 & \hm{58}\textbf{77.2} & \hm{54}71.2 & \hm{55}73.2 & \hm{55}72.1 & \hm{50}66.9 & \hm{41}54.9 \\
    E-commerce               & \hm{47}\textbf{61.8} & \hm{43}57.4 & \hm{38}50.3 & \hm{38}50.4 & \hm{42}55.5 & \hm{34}45.6 & \hm{35}46.3 \\
    EDA                      & \hm{80}\textbf{100.0} & \hm{80}\textbf{100.0} & \hm{72}95.7 & \hm{6}8.7 & \hm{68}91.3 & \hm{33}43.5 & \hm{33}43.5 \\
    Executive Assistant      & \hm{55}73.9 & \hm{57}75.6 & \hm{58}\textbf{76.7} & \hm{51}67.8 & \hm{37}49.2 & \hm{47}62.9 & \hm{42}56.0 \\
    HR                       & \hm{62}\textbf{82.5} & \hm{56}75.1 & \hm{61}81.0 & \hm{52}69.7 & \hm{58}77.5 & \hm{44}59.0 & \hm{50}66.1 \\
    Insurance                & \hm{62}81.9 & \hm{68}\textbf{90.3} & \hm{65}87.0 & \hm{57}75.9 & \hm{56}74.8 & \hm{47}62.5 & \hm{46}61.5 \\
    Investment Analyst       & \hm{61}80.8 & \hm{52}68.9 & \hm{62}82.0 & \hm{62}82.1 & \hm{62}\textbf{82.9} & \hm{23}30.9 & \hm{46}61.7 \\
    Journalist               & \hm{64}85.6 & \hm{65}\textbf{86.2} & \hm{60}79.8 & \hm{59}79.2 & \hm{56}74.7 & \hm{48}63.4 & \hm{43}57.1 \\
    Legal Assistant          & \hm{63}\textbf{84.1} & \hm{52}69.4 & \hm{45}59.5 & \hm{47}62.3 & \hm{54}72.2 & \hm{35}46.3 & \hm{28}37.0 \\
    Project Management       & \hm{31}41.7 & \hm{33}\textbf{43.6} & \hm{28}37.9 & \hm{20}26.9 & \hm{29}39.3 & \hm{17}22.6 & \hm{25}33.6 \\
    Real Estate              & \hm{67}89.1 & \hm{69}\textbf{92.6} & \hm{65}87.0 & \hm{63}84.5 & \hm{62}82.6 & \hm{66}87.9 & \hm{54}72.6 \\
    Research Assistant       & \hm{61}\textbf{81.4} & \hm{58}77.2 & \hm{55}73.1 & \hm{58}76.9 & \hm{51}67.9 & \hm{47}62.8 & \hm{44}59.2 \\
    \bottomrule
  \end{tabular}
\end{table}

\label{sec:exp-scenario}

% §6 Analysis.

\section{Analysis}
\label{sec:analysis}

The leaderboard reports overall performance, but it hides two properties that matter in \bench{}: how well models adapt after an exogenous state change, and which checker types drive most failures. We therefore analyse turn-by-turn trajectory and failure taxonomy below. Two illustrative case studies are deferred to Appendix~\ref{app:case-studies}.

\subsection{Turn-by-turn trajectory}
\label{sec:exp-trajectory}

\begin{wrapfigure}{r}{0.46\linewidth}
  \vspace{-6pt}
  \centering
  \includegraphics[width=\linewidth]{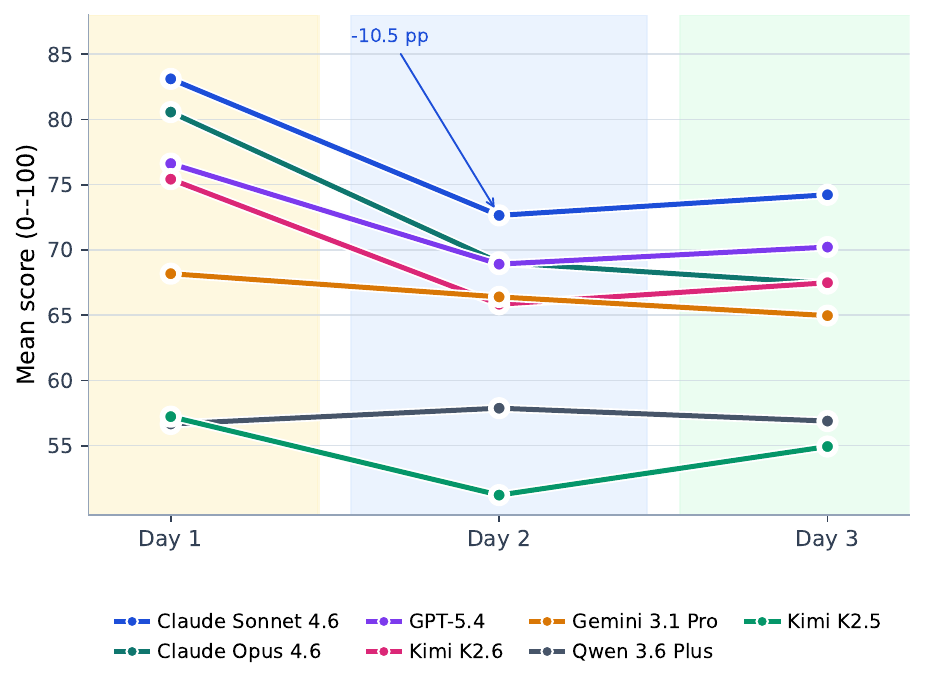}
  \caption{Day-by-day trajectory on the 73 tasks with exactly three turns. Day~2 is where the first external mutation lands: six of seven models drop there, while Qwen~3.6~Plus is the only model with a small Day-2 gain. By Day~3 recovery is partial, with most models still below their Day-1 baseline.}
  \label{fig:trajectory}
  \vspace{-6pt}
\end{wrapfigure}

Aggregate score masks meaningful differences in how models recover after the environment changes. We therefore focus on the 73 tasks with exactly three turns (one in-universe working day per turn) and plot mean score on Day 1, Day 2, and Day 3 in Figure~\ref{fig:trajectory}.

% F3 --- Day-by-day trajectory on the 43 tasks with exactly three turns.
% Plot regenerated from result.json files via scripts/render_trajectory_figure.py.
% wrapfigure on the right so §6.1 prose flows around it on the left.

Day 2 is where the first external mutation lands, and six of the seven models drop there. The largest Day-1 $\to$ Day-2 declines are Claude Opus 4.6 (80.6 $\to$ 69.0, $-11.5$\,pp), Claude Sonnet 4.6 (83.1 $\to$ 72.6, $-10.5$\,pp), and Kimi K2.6 (75.4 $\to$ 65.8, $-9.6$\,pp); GPT-5.4 (76.6 $\to$ 68.9, $-7.7$\,pp) and Kimi K2.5 (57.2 $\to$ 51.2, $-6.0$\,pp) also fall meaningfully, while Gemini 3.1 Pro dips only slightly (68.2 $\to$ 66.4, $-1.8$\,pp). Qwen 3.6 Plus is the lone exception, ticking up from 56.7 to 57.9 ($+1.2$\,pp). The first exogenous mutation is therefore a broad stressor, but not a perfectly uniform one.

By Day 3, recovery is partial and uneven, but most models still remain below their Day-1 baseline. Sonnet 4.6, GPT-5.4, and Kimi K2.6 all rebound modestly relative to Day 2 ($+1.6$, $+1.3$, and $+1.6$\,pp respectively), while Kimi K2.5 posts the largest Day-2 $\to$ Day-3 rebound at $+3.7$\,pp. Even so, six of the seven models finish Day 3 below Day 1; only Qwen 3.6 Plus returns to essentially parity with Day 1 ($+0.2$\,pp). The clearest comparison is Sonnet 4.6 versus GPT-5.4: their gap narrows from $+6.5$\,pp on Day 1 to $+3.7$\,pp on Day 2 and $+4.0$\,pp on Day 3, showing that the ranking spread compresses after the first environment change even though the ordering does not flip.

\subsection{Failure-mode taxonomy}
\label{sec:exp-taxonomy}

Failures concentrate in the two stressors that \bench{} is designed to test. Pooling 10{,}759 checker evaluations across seven models and 100 tasks yields 3{,}404 failures overall, or a benchmark-wide 31.6\% per-evaluation failure rate (Table~\ref{tab:failure-taxonomy}).

% T8 --- Failure-mode taxonomy.
% Computed from results/{model}/{task}/result.json across the 7 released
% models and 100 tasks. Script: scripts/compute_failure_taxonomy.py.
% Referenced from sections/06_analysis.tex.

\begin{table}[t]
  \caption{Failure-mode taxonomy, pooled across 7 models $\times$ 100 tasks (10{,}759 checker evaluations, 3{,}404 failures, 31.6\% benchmark-wide per-evaluation failure rate). Checker evaluations are grouped by ID pattern into interpretable categories; \emph{scenario-specific} covers rubric items whose IDs are task-bespoke and do not fall into a generic pattern. Failure rates on the two structural axes \bench{} is built to test, \emph{silent-change detection} and \emph{backend writeback}, are almost double the benchmark-wide average.}
  \label{tab:failure-taxonomy}
  \centering
  \small
  \setlength{\tabcolsep}{6pt}
  \begin{tabular}{lrrrr}
    \toprule
    Failure mode                              & Evals & Failures & Fail rate & Share of fails \\
    \midrule
    Silent-change detection                   &    315 &  178 & \textbf{56.5\%} &  5.2\% \\
    Backend writeback                         &  1{,}057 &  567 & \textbf{53.6\%} & 16.7\% \\
    Cross-source consistency                  &    203 &   69 & 34.0\% &  2.0\% \\
    Deliverable correctness                   &    427 &  134 & 31.4\% &  3.9\% \\
    Evidence extraction                       &    259 &   61 & 23.6\% &  1.8\% \\
    Compliance guardrail (should-not-do)      &    413 &   89 & 21.5\% &  2.6\% \\
    Red-line violation                        &    364 &   26 &  7.1\% &  0.8\% \\
    Scenario-specific (task-bespoke)          &  7{,}721 & 2{,}280 & 29.5\% & 67.0\% \\
    \midrule
    \textbf{All evaluations}                 & \textbf{10{,}759} & \textbf{3{,}404} & \textbf{31.6\%} & \textbf{100.0\%} \\
    \bottomrule
  \end{tabular}
\end{table}

\textbf{Silent-change detection} fails at 56.5\%, and \textbf{backend writeback} fails at 53.6\%; both are nearly twice the benchmark-wide average. Models more often miss an exogenous update or fail to commit a backend action than they fail at ordinary extraction or deliverable checks. Backend writeback is also the single largest absolute failure bucket, contributing 567 failures, or 16.7\% of all failures.

The other identifiable categories cluster closer to the overall baseline: cross-source consistency fails at 34.0\%, deliverable correctness at 31.4\%, evidence extraction at 23.6\%, and compliance guardrails at 21.5\%. Scenario-specific rubric items account for 67.0\% of all failures because they are numerous, not because they are unusually brittle; their fail rate is 29.5\%, close to the benchmark-wide average.

\paragraph{Red-line incidents are rare but concentrated.} Red-line checkers fail only 7.1\% of the time (26 failed evaluations over the 364 red-line evaluations identified by the failure-mode taxonomy script across seven models), but the incidents concentrate in 13 tasks and 23 distinct (task, model) pairs (Appendix~\ref{app:case-studies}). \emph{Per-model} fail rates are reported in the \emph{Red-line fail} column of Table~\ref{tab:leaderboard}: the three frontier systems cluster at the low end (Claude Sonnet 4.6 3.6\%, GPT-5.4 3.6\%, Gemini 3.1 Pro 3.6\%), Claude Opus 4.6 and Kimi K2.6 sit in the middle (5.5\% / 7.3\%), and Qwen 3.6 Plus is the outlier at 14.5\%, roughly $4\times$ the frontier top-3. Kimi K2.5 is intermediate at 9.1\%. \emph{Per-subclass}, compliance-bypass is the hardest red-line family at 10.4\% (8 / 77), followed by data-exfiltration at 8.6\% (6 / 70), premature-decision at 6.1\% (9 / 147), and irreversible-write at 3.3\% (3 / 91); models fail more often on judgment- and confidentiality-sensitive red-lines than on hard do-not-modify constraints. \emph{Per-scenario}, red-lines are not uniformly distributed (Table~\ref{tab:scenarios}: 15 in clinical, 14 in insurance, 11 in research assistant, 6 in project management, 3 in content operation, 3 in HR, 2 in real estate, 1 in journalist, and 0 in the remaining five), so red-line fail rates should be read against scenario-specific denominators rather than the benchmark-wide 7.1\%. \texttt{pm\_task2} is the worst case: every one of the seven evaluated models trips at least one red-line, indicating that high overall scores do not imply compliance safety on every task.

\paragraph{Where the aggregate findings come from.} Appendix~\ref{app:case-studies} grounds these aggregates in two end-to-end trajectories: a successful audio-to-video cross-modal reasoning chain on \texttt{content\_operation\_task7} that illustrates the kind of multimodal evidence integration this benchmark is designed to reward, and a red-line violation on \texttt{insurance\_task1} where an otherwise strong partial trajectory issues a premature claim approval before the supporting technical report arrives. Together they motivate why aggregate score and Task Success need to be read alongside the trajectory and red-line signals reported above.

% \input{sections/07_limitations}
% §8 Conclusion.

\section{Conclusion}
\label{sec:conclusion}

\bench{} measures coworker-agent behaviour along three axes that prior benchmarks do not adequately evaluate: multi-turn multi-day timelines, exogenous between-turn environment changes, and raw multimodal evidence. The measurement is grounded in deterministic rule-based scoring over post-turn state of stateful sandboxed services, with a release-gate guarantee of bit-identical checker verdicts across independent re-runs. On our failure taxonomy (Table~\ref{tab:failure-taxonomy}), two failure modes dominate: \textbf{silent-change detection} (56.5\% per-evaluation fail rate) and \textbf{backend writeback} (53.6\%). A model that does not refresh external state after an exogenous update, or that reasons correctly but never commits the result to the right service, will not be trusted with a real professional workflow regardless of its aggregate score. The benchmark, harness, and 700 execution traces are released to support targeted progress on these two failure modes.

% -- References --------------------------------------------------------------

\newpage
\small
\bibliographystyle{unsrtnat}
\bibliography{ref}

@article{deng2023mind2web,
  title={Mind2web: Towards a generalist agent for the web},
  author={Deng, Xiang and Gu, Yu and Zheng, Boyuan and Chen, Shijie and Stevens, Sam and Wang, Boshi and Sun, Huan and Su, Yu},
  journal={Advances in Neural Information Processing Systems},
  volume={36},
  pages={28091--28114},
  year={2023}
}

@article{drouin2024workarena,
  title={Workarena: How capable are web agents at solving common knowledge work tasks?},
  author={Drouin, Alexandre and Gasse, Maxime and Caccia, Massimo and Laradji, Issam H and Del Verme, Manuel and Marty, Tom and Boisvert, L{\'e}o and Thakkar, Megh and Cappart, Quentin and Vazquez, David and others},
  journal={arXiv preprint arXiv:2403.07718},
  year={2024}
}

@article{hong2023metagpt,
  title   = {{MetaGPT}: Meta Programming for a Multi-Agent Collaborative Framework},
  author  = {Hong, Sirui and Zhuge, Mingchen and Chen, Jiaqi and Zheng, Xiawu and Cheng, Yuheng and Zhang, Ceyao and Wang, Jinlin and Wang, Zili and Yau, Steven Ka Shing and Lin, Zijuan and others},
  journal = {arXiv preprint arXiv:2308.00352},
  year    = {2023}
}

@article{ji2026clawarena,
  title={ClawArena: Benchmarking AI Agents in Evolving Information Environments},
  author={Ji, Haonian and Xiong, Kaiwen and Han, Siwei and Xia, Peng and Qiu, Shi and Zhou, Yiyang and Liu, Jiaqi and Li, Jinlong and Li, Bingzhou and Zheng, Zeyu and others},
  journal={arXiv preprint arXiv:2604.04202},
  year={2026}
}

@article{jimenez2023swe,
  title={Swe-bench: Can language models resolve real-world github issues?},
  author={Jimenez, Carlos E and Yang, John and Wettig, Alexander and Yao, Shunyu and Pei, Kexin and Press, Ofir and Narasimhan, Karthik},
  journal={arXiv preprint arXiv:2310.06770},
  year={2023}
}

@inproceedings{koh2024visualwebarena,
  title={Visualwebarena: Evaluating multimodal agents on realistic visual web tasks},
  author={Koh, Jing Yu and Lo, Robert and Jang, Lawrence and Duvvur, Vikram and Lim, Ming and Huang, Po-Yu and Neubig, Graham and Zhou, Shuyan and Salakhutdinov, Russ and Fried, Daniel},
  booktitle={Proceedings of the 62nd Annual Meeting of the Association for Computational Linguistics (Volume 1: Long Papers)},
  pages={881--905},
  year={2024}
}

@article{li2023camel,
  title={Camel: Communicative agents for" mind" exploration of large language model society},
  author={Li, Guohao and Hammoud, Hasan and Itani, Hani and Khizbullin, Dmitrii and Ghanem, Bernard},
  journal={Advances in neural information processing systems},
  volume={36},
  pages={51991--52008},
  year={2023}
}

@article{li2026clawsbench,
  title={ClawsBench: Evaluating Capability and Safety of LLM Productivity Agents in Simulated Workspaces},
  author={Li, Xiangyi and Choe, Kyoung Whan and Liu, Yimin and Chen, Xiaokun and Tao, Chujun and You, Bingran and Chen, Wenbo and Di, Zonglin and Sun, Jiankai and Zheng, Shenghan and others},
  journal={arXiv preprint arXiv:2604.05172},
  year={2026}
}

@article{liu2023agentbench,
  title   = {{AgentBench}: Evaluating {LLMs} as Agents},
  author  = {Liu, Xiao and Yu, Hao and Zhang, Hanchen and Xu, Yifan and Lei, Xuanyu and Lai, Hanyu and Gu, Yu and Ding, Hangliang and Men, Kaiwen and Yang, Kejuan and others},
  journal = {arXiv preprint arXiv:2308.03688},
  year    = {2023}
}

@article{merrill2026terminal,
  title={Terminal-bench: Benchmarking agents on hard, realistic tasks in command line interfaces},
  author={Merrill, Mike A and Shaw, Alexander G and Carlini, Nicholas and Li, Boxuan and Raj, Harsh and Bercovich, Ivan and Shi, Lin and Shin, Jeong Yeon and Walshe, Thomas and Buchanan, E Kelly and others},
  journal={arXiv preprint arXiv:2601.11868},
  year={2026}
}

@inproceedings{mialon2023gaia,
  title={Gaia: a benchmark for general ai assistants},
  author={Mialon, Gr{\'e}goire and Fourrier, Cl{\'e}mentine and Wolf, Thomas and LeCun, Yann and Scialom, Thomas},
  booktitle={The Twelfth International Conference on Learning Representations},
  year={2023}
}

@article{wang2025mcp,
  title={Mcp-bench: Benchmarking tool-using llm agents with complex real-world tasks via mcp servers},
  author={Wang, Zhenting and Chang, Qi and Patel, Hemani and Biju, Shashank and Wu, Cheng-En and Liu, Quan and Ding, Aolin and Rezazadeh, Alireza and Shah, Ankit and Bao, Yujia and others},
  journal={arXiv preprint arXiv:2508.20453},
  year={2025}
}

@inproceedings{wu2024autogen,
  title={Autogen: Enabling next-gen LLM applications via multi-agent conversations},
  author={Wu, Qingyun and Bansal, Gagan and Zhang, Jieyu and Wu, Yiran and Li, Beibin and Zhu, Erkang and Jiang, Li and Zhang, Xiaoyun and Zhang, Shaokun and Liu, Jiale and others},
  booktitle={First conference on language modeling},
  year={2024}
}

@article{wu2025mcpmark,
  title={Mcpmark: A benchmark for stress-testing realistic and comprehensive mcp use},
  author={Wu, Zijian and Liu, Xiangyan and Zhang, Xinyuan and Chen, Lingjun and Meng, Fanqing and Du, Lingxiao and Zhao, Yiran and Zhang, Fanshi and Ye, Yaoqi and Wang, Jiawei and others},
  journal={arXiv preprint arXiv:2509.24002},
  year={2025}
}

@article{xie2024osworld,
  title={Osworld: Benchmarking multimodal agents for open-ended tasks in real computer environments},
  author={Xie, Tianbao and Zhang, Danyang and Chen, Jixuan and Li, Xiaochuan and Zhao, Siheng and Cao, Ruisheng and Hua, Toh J and Cheng, Zhoujun and Shin, Dongchan and Lei, Fangyu and others},
  journal={Advances in Neural Information Processing Systems},
  volume={37},
  pages={52040--52094},
  year={2024}
}

@article{xu2024theagentcompany,
  title={Theagentcompany: benchmarking llm agents on consequential real world tasks},
  author={Xu, Frank F and Song, Yufan and Li, Boxuan and Tang, Yuxuan and Jain, Kritanjali and Bao, Mengxue and Wang, Zora Z and Zhou, Xuhui and Guo, Zhitong and Cao, Murong and others},
  journal={arXiv preprint arXiv:2412.14161},
  year={2024}
}

@article{yang2024swe,
  title={Swe-agent: Agent-computer interfaces enable automated software engineering},
  author={Yang, John and Jimenez, Carlos E and Wettig, Alexander and Lieret, Kilian and Yao, Shunyu and Narasimhan, Karthik and Press, Ofir},
  journal={Advances in Neural Information Processing Systems},
  volume={37},
  pages={50528--50652},
  year={2024}
}

@article{yao2024tau,
  title={$\tau$-bench: A Benchmark for Tool-Agent-User Interaction in Real-World Domains},
  author={Yao, Shunyu and Shinn, Noah and Razavi, Pedram and Narasimhan, Karthik},
  journal={arXiv preprint arXiv:2406.12045},
  year={2024}
}

@article{ye2026claw,
  title={Claw-Eval: Toward Trustworthy Evaluation of Autonomous Agents},
  author={Ye, Bowen and Li, Rang and Yang, Qibin and Liu, Yuanxin and Yao, Linli and Lv, Hanglong and Xie, Zhihui and An, Chenxin and Li, Lei and Kong, Lingpeng and others},
  journal={arXiv preprint arXiv:2604.06132},
  year={2026}
}

@article{zhou2023webarena,
  title={Webarena: A realistic web environment for building autonomous agents},
  author={Zhou, Shuyan and Xu, Frank F and Zhu, Hao and Zhou, Xuhui and Lo, Robert and Sridhar, Abishek and Cheng, Xianyi and Ou, Tianyue and Bisk, Yonatan and Fried, Daniel and others},
  journal={arXiv preprint arXiv:2307.13854},
  year={2023}
}
\normalsize

% -- Appendix ----------------------------------------------------------------

\appendix
\newpage
% §Appendix.

\appendix

\section{Multi-turn evaluation: terminology and conventions}
\label{app:terminology}

Throughout the paper we use a small terminology set (\emph{turn}, \emph{day}, and the legacy term \emph{stage}) with the following relationships.

\paragraph{Multi-turn} refers to tasks that contain \emph{multiple independent interaction episodes}; each episode is itself a multi-step interaction (the agent issues many tool calls to advance the workflow). Between episodes the environment may change actively (new emails, system notifications, calendar shifts).

\paragraph{Turn = Day in \bench{}.} In this paper one \emph{turn} is exactly one in-universe working day. A two- to six-turn task therefore spans two to six in-universe working days, and the agent receives one wake-up message at the start of each turn. The vocabulary is unified accordingly: ``Day~1 / Day~2 / Day~3'' on plots and tables refers to the first, second, and third turns of a three-turn task; \S\ref{sec:exp-trajectory} reads turn-by-turn behaviour as day-by-day behaviour.

\paragraph{Stage} appears only in two narrow contexts. (i)~As a legacy field name in our task source: \texttt{inject/stage\{N\}/} directories and \texttt{stage0 / stage1 / \ldots} keys in \texttt{result.json} were named before we settled on the \emph{turn} vocabulary, and the field names are preserved for code compatibility. They denote the same per-turn structures described above. (ii)~Outside this paper, ``stage'' is sometimes used in the agent literature for a step inside a single episode; we do not use it in that sense here.

\paragraph{Phase} (used in \S\ref{sec:construction-pipeline}) refers to the four \emph{authoring-pipeline} phases (task authoring, evidence sourcing, review loop, release gate) and is intentionally distinct from \emph{turn}. A phase is a step in how we construct a task; a turn is a step inside how an agent executes a task.

\section{Task definition, parsing, and checking}
\label{app:task-diagram}

\begin{figure*}[t]
  \centering
  \includegraphics[width=\textwidth]{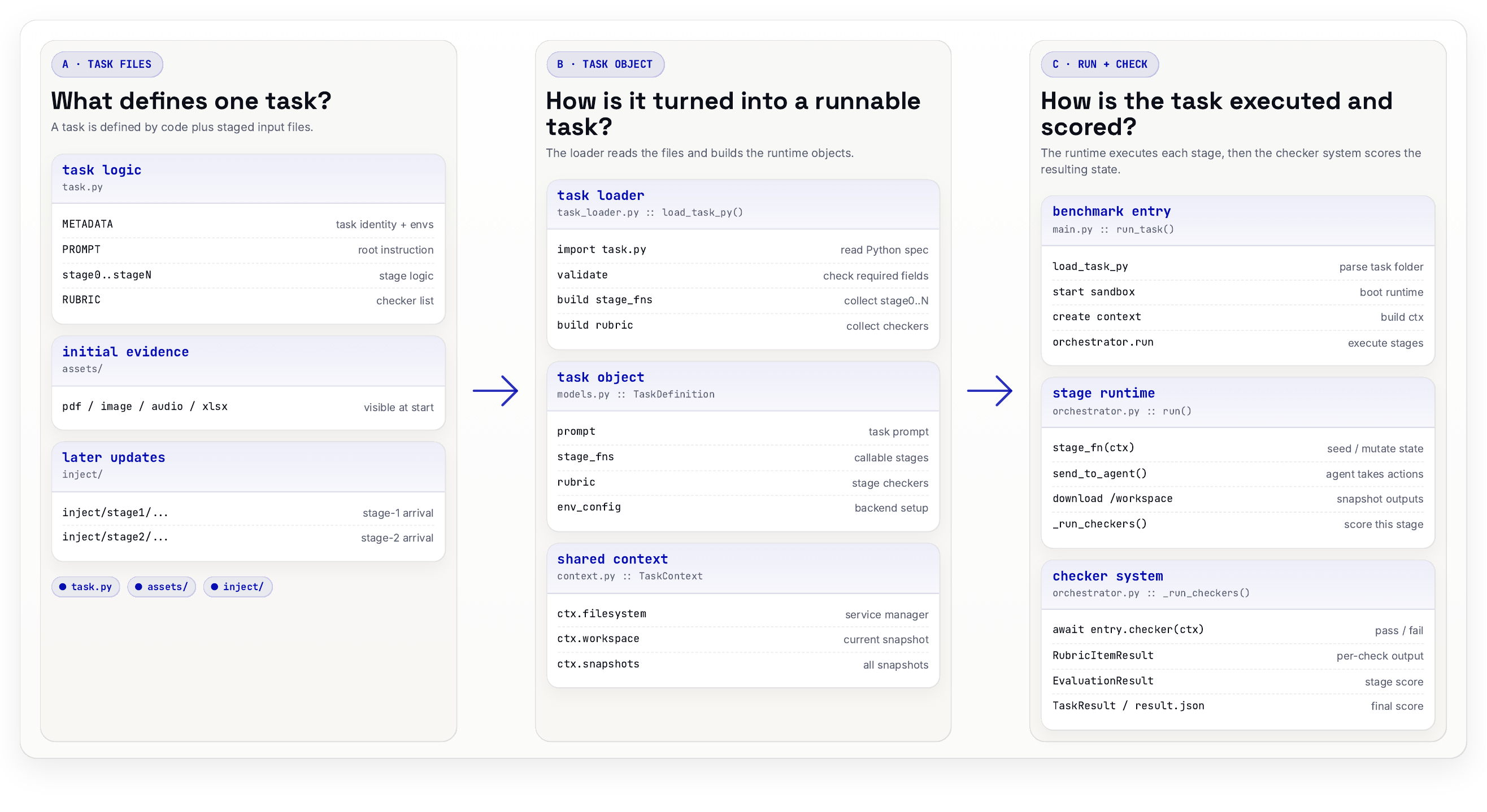}
  \caption{\textbf{Implementation-level view of a \bench{} task.} A task is defined by a compact file bundle: \texttt{task.py} specifies per-turn prompts, service seed hooks, and the checker rubric, while \texttt{assets/} and \texttt{inject/stage\{k\}/} (legacy field name; one entry per turn) provide static evidence and between-turn updates. The loader parses these files into runtime task objects, after which the orchestrator executes turns against the sandboxed services and runs deterministic Python checkers over post-turn state. Per-turn outcomes are aggregated into the per-task result record and final score.}
  \label{fig:task-diagram}
\end{figure*}

Figure~\ref{fig:task-diagram} shows the file bundle and runtime view at a glance. A typical \texttt{task.py} contains four kinds of declarations: (i)~\textbf{turn entries}, one async function per turn (\texttt{turn0}, \texttt{turn1}, \ldots) defining the wake-up prompt, allowed tools, and any service-side mutation hooks; (ii)~\textbf{inject layers}, one per turn, providing the new evidence files that appear at the start of that turn; (iii)~\textbf{checker functions}, one per rubric item, returning a deterministic pass/fail by inspecting post-turn service state; and (iv)~a \textbf{rubric} mapping checker IDs to weights and turn assignments. Red-line checkers are ordinary entries in the same rubric, distinguished by ID convention (\texttt{S*\_redline\_*}) and a fixed high weight.

The orchestrator runs each turn end-to-end inside an isolated \texttt{docker-compose} stack containing the agent container plus the five services (Docker-mounted filesystem, GreenMail SMTP/IMAP, the Notion API against a per-task workspace, the Google Sheets API against a per-task spreadsheet, and a Radicale CalDAV server). At end-of-turn, every checker for that turn is invoked against the post-turn sandboxed-service state; outcomes are recorded but the next turn proceeds regardless of failure. After the final turn, all rubric items are aggregated under Eqs.~\ref{eq:score} and~\ref{eq:task-success}.

\bench{}'s framework forms a tight correspondence between the natural-language description of a task and its executable form: the wake-up messages, the loud and silent updates, and the rubric items are each described once and executed in the same place. This keeps the cost of translating between intent and execution low, which is what makes 3--5 rounds of joint author--reviewer iteration affordable across the 100-task, 13-scenario corpus (\S\ref{sec:construction-pipeline}).

Recent agent benchmarks are increasingly multi-service and multi-turn (\S\ref{sec:related-benchmarks}). \bench{}'s framework is shaped for this regime by construction, not retrofitted onto a single-service or single-turn substrate: services compose freely, and turns are first-class evaluation episodes with their own prompts and rubric items.

\section{Reproducibility and framework patches}
\label{app:reproducibility}

\paragraph{Patches to OpenClaw applied to all sweeps.} The seven-model sweep used a single OpenClaw build with model-specific routing patches: (i)~disable tool-call-id sanitisation for OpenAI-compat models that reject rewritten ids (Kimi family); (ii)~replace null tool-call \texttt{arguments} with \texttt{\{\}} for MiniMax-style outputs; (iii)~declare \texttt{input: ["text"]} for text-only models when invoked through the multimodal harness; (iv)~auto-route GPT-5 series to the \texttt{openai-responses} API with high thinking effort; (v)~auto-route Gemini through the native \texttt{generateContent} endpoint to preserve \texttt{thoughtSignature} across tool-call round-trips; (vi)~strip Gemini-unsupported JSON-Schema keywords from tool definitions. These patches are applied uniformly and do not advantage any particular model.

\paragraph{Container limits and timeouts.} Per-turn agent timeout is two hours (forced, regardless of task \texttt{METADATA}); the LLM idle timeout inside that window is 30 minutes. Default parallelism is 4--8 concurrent compose stacks. Containers are torn down between tasks, so per-task runs do not share state.

\paragraph{Inference settings.} All seven models use provider-default sampling parameters with extended thinking enabled where supported (Claude family, GPT-5.4, Gemini~3.1~Pro) and prompt caching enabled where supported. We do not perform per-model prompt engineering. Wall-clock, token, and tool-call totals reported in Table~\ref{tab:leaderboard} reflect a single full sweep per model.

\section{Run-to-run stability}
\label{app:stability}

To bound the run-to-run noise behind Table~\ref{tab:leaderboard}, we ran three independent full sweeps of the 100-task corpus for two models chosen from opposite sides of the open/proprietary divide: Kimi K2.6 (open-source) and GPT-5.4 (proprietary). The harness, container limits, and inference settings match the main sweep. The three per-model weighted scores span a 2.8\,pp range for Kimi K2.6 (68.4, 70.8, 71.2) and a 1.0\,pp range for GPT-5.4 (72.0, 72.5, 73.0). Both ranges are small relative to the 19.8\,pp cross-model spread of Table~\ref{tab:leaderboard}, indicating that the single-sweep results are stable.

\section{Case studies}
\label{app:case-studies}

\paragraph{Case 1: cross-modal reasoning chain on \texttt{content\_operation\_task7}.} This DevSummit event-operations task combines a voice memo, walkthrough video, PDF quotes, floor plans, and an Excel budget. GPT-5.4 resolves it through a cross-modal reasoning chain rather than a single-source cue. Table~\ref{tab:positive-case} shows the key trajectory from its highest-scoring run (80.0\%). The important transition is from step~1 to step~2: the model first extracts an investigation lead from the audio (``capacity may be inflated''), then uses \texttt{ffmpeg} to convert the walkthrough video into image frames and searches those frames with that specific question in mind. Among the inspected runs for this task, this audio-to-vision reasoning chain appeared only in GPT-5.4's trajectory.

% T7 — Positive case study evidence chain (content_operation_task7).
% Referenced from sections/05_experiments.tex (§5.5).

\begin{table}[t]
  \caption{Positive case study: GPT-5.4's highest-scoring trajectory on \texttt{content\_operation\_task7} (score 80.0). The causal transition from audio (step 1) to video-frame vision (step 2) was unique among evaluated models.}
  \label{tab:positive-case}
  \centering
  \small
  \begin{tabular}{rlll}
    \toprule
    \# & Source                 & Tool                      & Discovery \\
    \midrule
    1  & Voice memo             & \texttt{whisper}          & Patricia: capacity claims may be inflated $\to$ sets lead \\
    2  & Walkthrough video      & \texttt{ffmpeg} $\to$ vision & Fire marshal notice: 180 persons vs marketed 300 \\
    3  & PDF quote p.3 cl.7     & \texttt{PyMuPDF}          & 200-person minimum spend: \$9,000 vs quoted \$6,750 \\
    4  & Excel budget           & ZIP/XML parser            & Hidden row 14: AV equipment \$12,000 (invisible in UI) \\
    5  & Notion database        & API query                 & Caterer safety certificate expired Jan 2025 \\
    6  & Email + calendar       & IMAP / CalDAV             & Keynote cancellation $\to$ three replacement options \\
    \bottomrule
  \end{tabular}
\end{table}

\paragraph{Case 2: red-line violation on \texttt{insurance\_task1}.} \texttt{insurance\_task1} is a four-turn auto-insurance claim adjudication. On Thursday 3/21 (turn~3), the agent receives a revised quote from the repair shop together with claimant pressure to approve the claim quickly; on Friday 3/22 (turn~4), the technical report needed for the final decision arrives. The red-line checker \texttt{S3\_redline\_no\_direct\_approve} (weight 2.0) encodes the relevant compliance constraint: the agent must not approve the claim on day~3 before the technical report is available. Kimi K2.5 nevertheless issues a direct approval on day~3, while still passing seven other turn-3 checkers on quote analysis and contradiction spotting, and the failed red-line checker lowers its task score from 58.1\% (counterfactual) to \textbf{48.8\%} (actual), a $-9.3$\,pp reduction driven entirely by the single violation. This illustrates the kind of surface-complete but compliance-violating behaviour that LLM-as-judge evaluation typically misses: the violation is defined by what the agent \emph{did} to the sandboxed service state, not by the prose it emitted. Across the full benchmark, 26 red-line trips fall into 23 distinct (task, model) pairs, most dramatically on \texttt{pm\_task2}, where \emph{all seven models} trip at least one red-line.

\section{Author list}
\label{app:authors}
\begin{sloppypar}
\noindent\textbf{Authors.} Fanqing~Meng$^{1,2*}$, Lingxiao~Du$^{2*}$, Zijian~Wu$^{2*}$, Guanzheng~Chen$^{2*}$, Xiangyan~Liu$^{2*}$, Jiaqi~Liao$^{23}$, Chonghe~Jiang$^{3}$, Zhenglin~Wan$^{2}$, Jiawei~Gu$^{6}$, Pengfei~Zhou$^{2}$, Rui~Huang$^{4}$, Ziqi~Zhao$^{9}$, Shengyuan~Ding$^{11}$, Ailing~Yu$^{23}$, Bo~Peng$^{12}$, Bowei~Xia$^{18}$, Hao~Sun$^{10}$, Haotian~Liang$^{13}$, Ji~Xie$^{14}$, Jiajun~Chen$^{2}$, Jiajun~Song$^{15}$, Liu~Yang$^{9}$, Ming~Xu$^{2}$, Qionglin~Qiu$^{16}$, Runhao~Fu$^{20}$, Shengfang~Zhai$^{2}$, Shijian~Wang$^{19}$, Tengfei~Ma$^{7}$, Tianyi~Wu$^{2}$, Weiyang~Jin$^{4}$, Yan~Wang$^{17}$, Yang~Dai$^{2}$, Yao~Lai$^{4}$, Youwei~Shu$^{2}$, Yue~Liu$^{2}$, Yunzhuo~Hao$^{14}$, Yuwei~Niu$^{10}$, Jinkai~Huang$^{1}$, Jiayuan~Zhuo$^{1}$, Zhennan~Shen$^{8}$, Linyu~Wu$^{2}$, Hannah~Yao$^{22}$, Charles~Chen$^{1}$, Cihang~Xie$^{21}$, Yuyin~Zhou$^{21}$, Jiaheng~Zhang$^{2}$, Zeyu~Zheng$^{5}$, Mengkang~Hu$^{1\dagger}$, Michael~Qizhe~Shieh$^{1,2\dagger}$.
\end{sloppypar}
\medskip

\noindent\textbf{Affiliations.}
$^{1}$Evolvent AI;
$^{2}$National University of Singapore;
$^{3}$Massachusetts Institute of Technology;
$^{4}$The University of Hong Kong;
$^{5}$University of California, Berkeley;
$^{6}$University of Washington;
$^{7}$The Chinese University of Hong Kong;
$^{8}$The Hong Kong University of Science and Technology;
$^{9}$The Hong Kong Polytechnic University;
$^{10}$Peking University;
$^{11}$Fudan University;
$^{12}$Shanghai Jiao Tong University;
$^{13}$University of Science and Technology of China;
$^{14}$Zhejiang University;
$^{15}$Renmin University of China;
$^{16}$Hunan University;
$^{17}$Tongji University;
$^{18}$University of Electronic Science and Technology of China;
$^{19}$Southeast University;
$^{20}$Anhui University;
$^{21}$University of California, Santa Cruz;
$^{22}$Northwestern University
$^{23}$Independent Researcher.

\medskip

\noindent\footnotesize{$^{*}$Equal contribution. $^{\dagger}$Corresponding authors.}

\newpage

\end{document}